\documentclass[runningheads]{llncs}
\usepackage{graphicx}
\usepackage{adjustbox}
\usepackage{textcomp}
\usepackage{booktabs}
\usepackage[textsize=scriptsize,textwidth=1cm]{todonotes}

\begin{document}
\title{Towards a Grounded Dialog Model for Explainable Artificial Intelligence}
%
%
\author{Prashan Madumal \and
Tim Miller \and
Frank Vetere \and
Liz Sonenberg
}
\authorrunning{P. Madumal et al.}
%
\institute{School of Computing and Information Systems\\
University of Melbourne, Parkville, Australia.\\
\email{pmathugama@student.unimelb.edu.au}\\
\email{\{tmiller,f.vetere,l.sonenberg\}@unimelb.edu.au}}
\maketitle              
\begin{abstract}
To generate trust with their users, Explainable Artificial Intelligence (XAI) systems need to include an explanation model that can communicate the internal decisions, behaviours and actions to the interacting humans. Successful explanation involves both cognitive and social processes. In this paper we focus on the challenge of meaningful interaction between an explainer and an explainee and investigate the structural aspects of an explanation in order to propose a human explanation dialog model. We follow a bottom-up approach to derive the model by analysing transcripts of 398 different explanation dialog types. We use grounded theory to code and identify key components of which an explanation dialog consists. We carry out further analysis to identify the relationships between components and sequences and cycles that occur in a dialog. We present a generalized state model obtained by the analysis and compare it with an existing conceptual dialog model of explanation. 

\end{abstract}
\section{Introduction}

Explanation is important to artificial intelligence (AI) systems that aim to be transparent about their actions, behaviours and decisions. This is especially true in scenarios where a human needs to take critical decisions based on outcome of an AI system. A proper explanation model aided with argumentation can promote trust humans have about the system, allowing better cooperation~\cite{Reference16} in their interactions, in that humans have to reason about the extent to which, if at all, they should trust the provider of the explanation.

	As Miller~\cite[pg 10]{Reference13} notes, the process of Explanation is inherently socio-cognitive as it involves two processes: (a) a \emph{Cognitive process}, namely the process of determining an explanation for a given event, called the \emph{explanandum}, in which the causes for the event are identified and a subset of these causes is selected as the explanation (or \emph{explanans}); and (b) the \emph{Social process} of transferring knowledge between explainer and explainee, generally an interaction between a group of people, in which the goal is that the explainee has enough information to understand the causes of the event.
    
However, much research and practice in explainable AI uses the researchers' intuitions of what constitutes a `good' explanation rather basing the approach on a strong understanding of how people define, generate, select, evaluate, and present explanations~\cite{Reference13,miller2017explainable}. Most modern work on Explainable AI, such as in autonomous agents~\cite{winikoff2017debugging,broekens2010you,Reference20,Reference21} and interpretable machine learning~\cite{Reference19}, does not discuss the interaction aspect of the explanations. Lack of a general dialog model of explanation that takes into account the end user can be attributed as one of shortcomings of existing explainable AI systems. Although there are existing conceptual explanation dialog models that try to emulate the structure and sequence of a natural explanation~\cite{Reference6,Reference4}, we propose that improvements will come from further generalization. 

Explanation naturally occurs as a continuous interaction which gives the interacting party the ability to question and interrogate the explanations. This enables the explainee to clear doubts about the given explanation by further interrogations and user-driven questions. Further, the explainee can express contrasting views about the explanation that can set the premise for an argumentation dialog. This type of iterative explanation has the ability to provide more rich and satisfactory explanations as opposed to singular explanation presentations.

Understanding how humans engage in conversational explanation is a prerequisite to building an explanation model as noted by Hilton~\cite{Reference22}. De Graaf~\cite{Reference3} note that humans attribute human traits, such as beliefs, desires, and intentions, to intelligent agents, and it is thus a small step to assume that people will explain agent behaviour using human frameworks of explanation. AI explanation models with designs that are influenced by human explanation models should be able to provide more intuitive explanations to humans and therefore be more likely to be accepted. We suggest it is easier for the AI to emulate human explanations rather than expecting humans to adapt to a novel and unfamiliar explanation model. While there are mature existing models for explanation dialogs~\cite{Reference4,Reference15}, these are idealised conceptual models that are not grounded on or validated by data, and seem to lack iterative features like cyclic dialogs.

In this paper our goal is to introduce a Human explanation model that is based on conversational data. We follow a bottom up approach and aim to provide an explanation dialog model that is grounded on data from different types of explanations in conversations. We derive our model by analysing 398 explanation dialogs across six different types of dialogs. Frequency, sequence and relationships between the basic components of an explanation dialog were obtained and analyzed in the study.  We believe by following a data driven approach to formulate the model, a more generalized and accurate model can be developed that can define the structure and the sequence of an explanation dialog.

This paper is structured as follows, we discuss related work regarding explanation in AI and explanation dialog models, then we explain the methodology of the study and collection of data and its properties. We analyse and evaluate the data in section 4, identifying key components of an explanation dialog and gaining insight to the relationships and sequence of these components. We also develop the explanation dialog model based on the analysis and compare it with a similar model by Walton~\cite{Reference8}.
We end the paper by discussing the model with its contribution and significance in socio-cognitive systems.

\section{Related Work}

Explaining decisions of intelligent systems has been a topic of interest since the era of expert systems, e.g.~\cite{Reference1,KassFinin88}. Early work focussed particularly on the explanation’s content, responsiveness and the human-computer interface through which the explanation was delivered. 
Kass and Finin~\cite{KassFinin88} and Moore and Paris~\cite{Reference17} discussed the requirements a good explanation facility should have, 
including characteristics like ``Naturalness'', and pointed to the critical role of user models in explanation generation.
Cawsey's~\cite{Reference2} EDGE system also focused on user interaction and user knowledge. These were used to update the system through interaction. So, from the very early days, both the  cognitive and social attributes associated with an agent's awareness of other actors, and capability to interaction with them, has been recognized as an essential feature of explanation research. However, limited progress has been made. Indeed recently, de Graaf and Malle~\cite{Reference3} still find the need to emphasize the importance of understanding how humans respond to Autonomous Intelligent Systems (AIS). They further note how Humans will expect a familiar way of communication from AIS systems when providing explanations.

Castelfranchi~\cite{Castelfranchi:1998:PTM:551984.852234} argues the importance of `trust' in a social setting in Multi-Agent Systems (MAS). MAS often have socialites such as Cooperation, collaboration and team work which are influenced by trust. Castelfranchi further explains the components of social trust which are rooted in delegation. Persistence belief and self-confidence belief~\cite{Castelfranchi:1998:PTM:551984.852234} in weak delegation can arguably fulfilled by providing satisfactory explanations about agents intentions and behaviors.

To accommodate the communication aspects of explanations, several dialog models have been proposed.  Walton~\cite{Reference4,Reference15} introduces a shift model that has two distinct dialogs: an explanation dialog and an examination dialog, where the latter is used to evaluate the success of an explanation. Walton draws from the work of Memory Organizing Packages (MOP)~\cite{Reference18} and case-based reasoning to build the routines of the explanation dialog models. Walton's dialog model has three stages namely the opening stage, an argumentation stage and a closing stage~\cite{Reference4}. Walton suggests an examination dialog with two rules as the closing stage. These rules are governed by the explainee, which corresponds to the understanding of an explanation~\cite{Reference5}. This sets the premise for the examination dialog of an explanation and the shift between explanation and examination to determine the success of an explanation~\cite{Reference15}. 

A formal dialogical system of explanation is also proposed by Walton~\cite{Reference5} that has three types of conditions: Dialog conditions; Understanding conditions; and the success conditions that constitutes an explanation. Arioua and Croitoru~\cite{Reference6} formalized and extended Walton’s explanatory CE dialectical system by incorporating Prakken’s~\cite{Reference7} framework of dialog formalisation.

Argumentation also comes into to play in explanation dialogs. Walton~\cite{Reference8} introduced  a dialog system for argumentation and explanation that consists of communication language that defines the speech acts and protocols which allow transitions in the dialog. This allows the explainee to challenge and interrogate the given explanations to gain further understanding. Villata et al.~\cite{Reference16} focus on modelling information sources to be suited in an argumentation framework, and introduce a socio-cognitive model of trust to support judgements about trustworthiness. 

\section{Research Design and Methodology}

To address the lack of an explanation dialog model that is based on and evaluated through conversation data, we opt to use a data-driven approach. This study consists of a data selection and gathering phase, data analysis phase and a model development phase.

We designed the study as a bottom up approach to develop an explanation dialog model. We aimed to gain insights into three areas: 1. Key components that makeup an explanation dialog; 2. Relationships that exist within those components; and 3. Component sequences that occur in an explanation dialog and cycles. 

\subsection{Research design}
We formulate our design based on an inductive approach. We use grounded theory~\cite{Reference9} as the methodology to conceptualize and derive models of explanation. The key goal of using grounded theory as opposed to using a hypothetico-deductive approach is to formalize an explanation dialog model that is grounded on actual conversation data of various types rather than a purely conceptual model.

The study is divided into three distinct stages, based on grounded theory. The first stage consists of coding~\cite{Reference9} and theorizing, where small chunks of data are taken, named and marked according to the concepts they might hold. For example a segment of a paragraph in an interview transcript can be identified as an `Explanation' and another segment can be identified as a `Why question'. This process is repeated until the whole data set is coded. The second stage involves categorizing, where similar codes and concepts are grouped together by identifying their relationship with each other. The third stage involves deriving a theoretical model from the codes, categories and their relationship.

\subsection{Data}

We collect data from six different data sources that have six different types of explanation dialogs. Table \ref{tab1} shows the explanation dialog types, explanation dialogs that are in each type and number of transcripts. We gathered and coded a total of 398 explanation dialogs from all of the data sources. All the data sources are text based, where some of them are transcribed from voice and video-based interviews. Data sources consist of Human-Human conversations and Human-Agent conversations\footnote{\label{note1}Links to all  data sources (including transcripts) can be found at \url{https://explanationdialogs.azurewebsites.net}}. We collected Human-Agent conversations to analyze if there are significant differences in the way humans carry out the explanation dialog when they knew the interacting party was an agent.

\begin{table}[!h]
\caption{Coded data description.}
\label{tab1}

\begin{center}
\begin{tabular}{lrrlr}
\toprule
Explanation Dialog Type &  \# Dialogs & ~~\# Transcripts  \\
\midrule
1. Human-Human static explainee &  88 & 2 \\
2. Human-Human static explainer & 30 & 3 \\
3. Human-Explainer agent & 68 & 4 \\
4. Human-Explainee agent & 17 & 1 \\
5. Human-Human QnA & 50  & 5 \\
6. Human-Human multiple explainee & 145 & 5 \\
\bottomrule
\end{tabular}

\end{center}
\vskip -0.1in
\end{table}

Data source selection was done to encompass different combinations and frequencies an explainee and explainer included in the explanation dialog. These combinations are given in Table \ref{tab3}. We diversify the dataset by including data sources of different mediums such as verbal based and text based.

\begin{table}[!h]
\caption{Explanation dialog type description.}
\label{tab3}
\begin{center}

\begin{tabular}{llll}
\toprule
Participants &  Number & Medium~ & Data source  \\
\midrule
1. Human-Human~  &  One-to-one & Verbal & Journalist Interview transcripts \\
2. Human-Human  & One-to-one & Verbal & Journalist Interview transcripts \\
3. Human-Agent & One-to-one & Text & Chatbot conversation transcripts \\
4. Agent-Human & One-to-one & Text & Chatbot conversation transcripts \\
5. Human-Human & Many-to-many~ & Text  & Reddit AMA records \\
6. Human-Human & One-to-many & Verbal & Supreme court transcripts \\
\bottomrule
\end{tabular}

\end{center}
\vskip -0.2in
\end{table}

Table \ref{tab2} presents the coding, categories and their definition. We identify why, how and what questions as questions that ask contrastive explanations, questions that ask explanations of causal chains and questions that ask causality explanations in that order.

\begin{table}[!h]
\caption{Code description.}
\label{tab2}
\vskip 0.01in
\begin{center}
\begin{small}

\begin{tabular}{llp{5.5cm}}
\toprule
Code &  Category & Description  \\
\midrule
QE start &   Dialog & Explanation dialog start\\
QE end &  Dialog & Explanation dialog end\\
How & Question Type~ & How questions\\
Why & Question Type & Why questions\\
What & Question Type  & What questions \\
Explanation & Explanation & Explanation given for questions\\
Explainee Affirmation &  Explanation & Explainee acknowledges explanation 
 \\
Explainer Affirmation & Explanation &  Explainer acknowledges explainee's acknowledgment\\
Question context & Information & Background to the question provided by the explainee\\
Preconception & Information & Preconceived idea that the explainee has about some fact\\
Counterfactual case & Information & Counterfactual case of the how/why question  \\
Argument & Argumentation  & Argument presented by explainee or explainer  \\
Argument-s & Argumentation & An argument that starts the Dialog\\
Argument-a &  Argumentation & Argument Affirmation by explainee or explainer\\
Argument-c & Argumentation & Counter argument\\
Argument-contrast case & Argumentation & Argumentation contrast case
\\
Explainer Return question~ & Questions & Clarification question by explainer\\
Explainee Return question & Questions  & Follow up question asked by explainee
 \\
\bottomrule
\end{tabular}

\end{small}
\end{center}
\vskip -0.3in
\end{table}


\textbf{Explanation dialog type 1:} Same Human explainee with different Human explainers in each data source. Data sources for this type is journalist interviews where the explainee is the journalist.

\textbf{Explanation dialog type 2:} Same Human explainer with different Human explainees. Data source consist of journalists interviewing Malcolm Turnbull (Current Australian Prime Minister).

\textbf{Explanation dialog type 3:} Same Agent explainer with different Human explainees. Data source is 2016 Loebner Prize~\cite{Reference10} judge transcripts.

\textbf{Explanation dialog type 4:} Same Agent explainee with different Human explainers. Data source consists of transcript of conversation of Eliza chatbot~\cite{Reference10}.

\textbf{Explanation dialog type 5:} Multiple Human explainers and Multiple Human explainees. Data source is from reddit ask me anything (ama)~\cite{Reference11} records.

\textbf{Explanation dialog type 6:} Multiple Human explainees with a Human explainer. Data source consists of cases transcribed in supreme court of United States of America.

\section{Results}

In this section, we present the model resulting from our study, compare it to an existing conceptual model, and analyse some patterns of interaction that we observed.

\subsection{Explanation Dialog Model}

\begin{figure}[!t]
\includegraphics[trim={0 0cm 0 0cm}, width=\columnwidth]{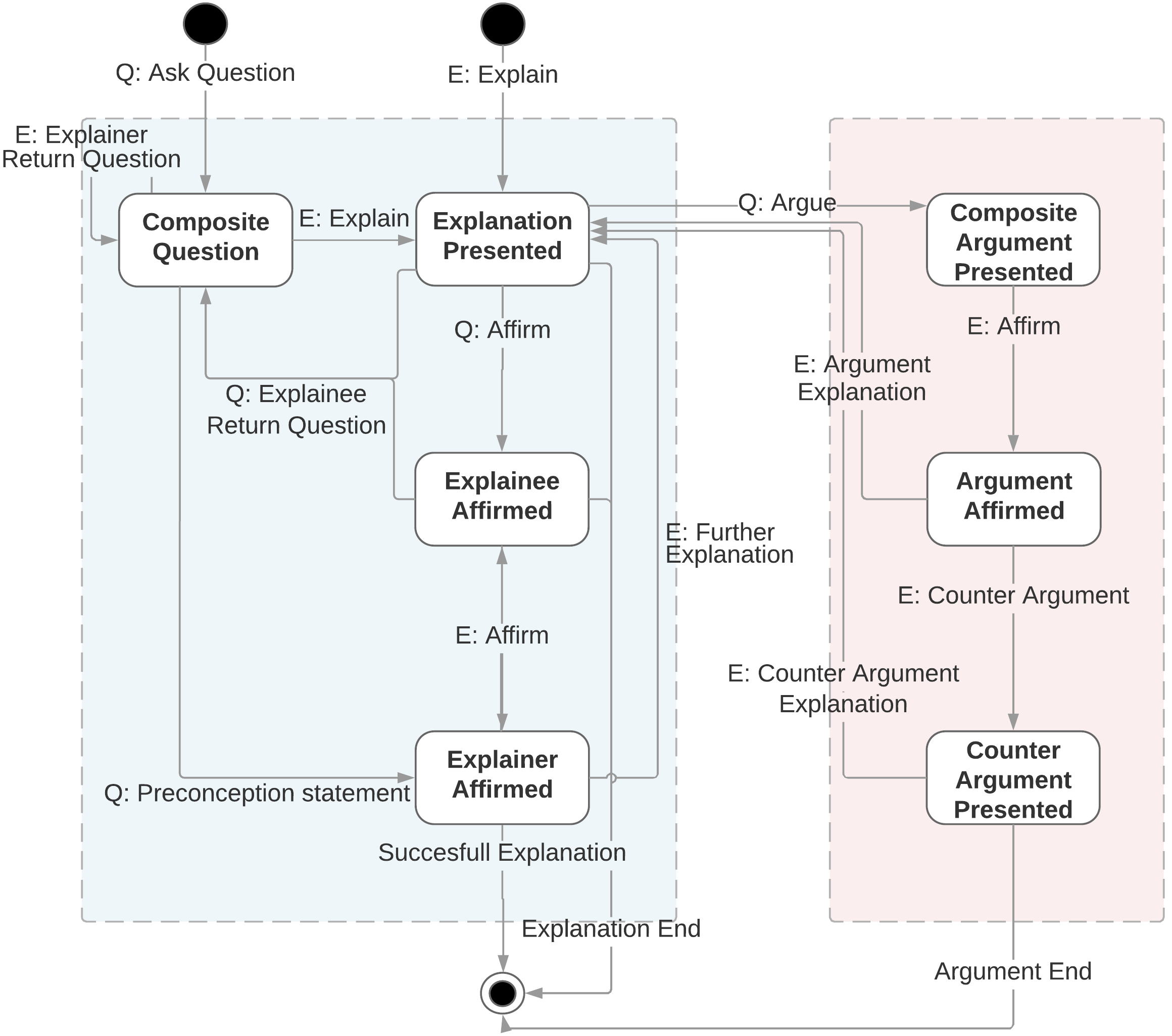}
\caption{Explanation Dialog Model} \label{fig5}
\end{figure}

We present our explanation dialog model derived from the study. Figure \ref{fig5} depicts the derived explanation dialog model as a state diagram where the sequence of each component is preserved. The labels `Q' and `E' refer to the Questioner (the explainee) and the Explainer respectively.


While most codes are directly transferred to the model as states and state transitions, codes that belonged to information category are embedded in different states. These are: 1. “Question Context”, “Preconception” and “Counterfactual Case” embedded in the “Composite Question” state; and 2. “Argumentation Contrast Case” embedded in the “Argumentation Dialog Initiated” state. These embedded components can potentially be used to decode the question or the argument by systems that uses the model.


We identify two loosely coupled sections of the model: the \emph{Explanation Dialog} and the \emph{Argumentation Dialog}. These two sections can occur in any order, frequencies and cycles. An explanation dialog can either be initiated by an argument or a question. An argument can occur at any point in the dialog time-line after an explanation was given, which will then continue on to an argumentation dialog. The dialog can then end in the argumentation dialog or can again be switched to the explanation dialog as shown in Figure \ref{fig5}. A single dialog can contain many argumentation dialogs and can go around the explanation dialog loop several times, which will switch in between them according to the flow of the dialog. Note that a loop in the explanation dialog implies that the ongoing explanation is related to the same original question. We coded explanation dialogs to end when a new topic was raised in a question. Questions that ask for follow-up (Return Questions) were coded when the questions were clearly identifiable as requesting more information about the given explanation.  

A typical example would be, an explainee asking a question, receiving a reply, presenting and argument of the explanation, explainer acknowledging the argument, explainer agreeing to the argument. This scenario can be described in states as Start \textrightarrow Composite Question \textrightarrow Explanation Presented \textrightarrow Explainee Affirmed \textrightarrow Argumentation Dialog Initiated \textrightarrow Composite Argument Presented \textrightarrow Argument Affirmed \textrightarrow End.

\subsubsection{Model Comparison}

\begin{figure}[!h]
\includegraphics[width=\columnwidth]{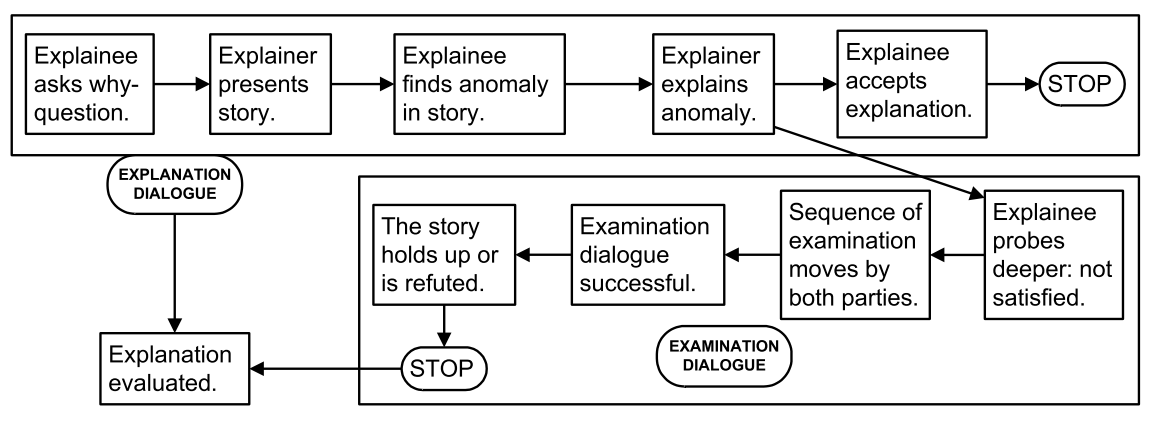}
\caption{Argumentation and explanation in dialogue~\cite{Reference8} } \label{fig6}
\end{figure}

We compare the developed explanation dialog model, which also contains an argumentation sub-dialog, by Walton~\cite{Reference8}. Walton proposed the model shown in Figure~\ref{fig6}, which consists of 10 components. This model focus on combining explanation and examination dialogs with argumentation.  A similar shift between explanation and argumentation/examination can be seen between our model and Walton's. According to the data sources, argumentation is a frequently present component of an explanation dialog, which is depicted by the Explainee probing component in Walton's Model. The basic flow of explanation is the same between the two models, but the models differ in two key ways. First, is the lack of \emph{examination} dialog shift in our model. Although we did not derive an examination dialog, a similar shift of dialog can be seen with respect to affirmation states. That is, our `examination' is simply the explainee affirming that they have understood the explanation.  Second is Walton’s model~\cite{Reference8} focus on the evaluation of the successfulness of an explanation in the form of examination dialog whereas our model focus on delivering an explanation in a natural sequence without an explicit form of explanation evaluation.


Thus, we can see similarities between Walton's conceptual model and our data-driven model. The differences between the two are at a more detailed level than at the high-level, and we attribute these differences to the grounded nature of our study. While Walton proposes an idealised model of explanation, we assert that our model captures the subtleties that would be required to build a natural dialog for human-agent explanation.

\subsection{Analysis and Evaluation}

We focus our analysis on three areas: 1. The key components of an Explanation Dialog; 2. Relationships between these components and their variations between different dialog types; and 3. The sequence of components that can successfully carry out an explanation dialog. We also evaluate the frequencies of certain component sequences to occur across different types of explanation dialogs. 

\subsubsection{Code Frequency Analysis}

\begin{figure}[!h]
\includegraphics[width=\textwidth]{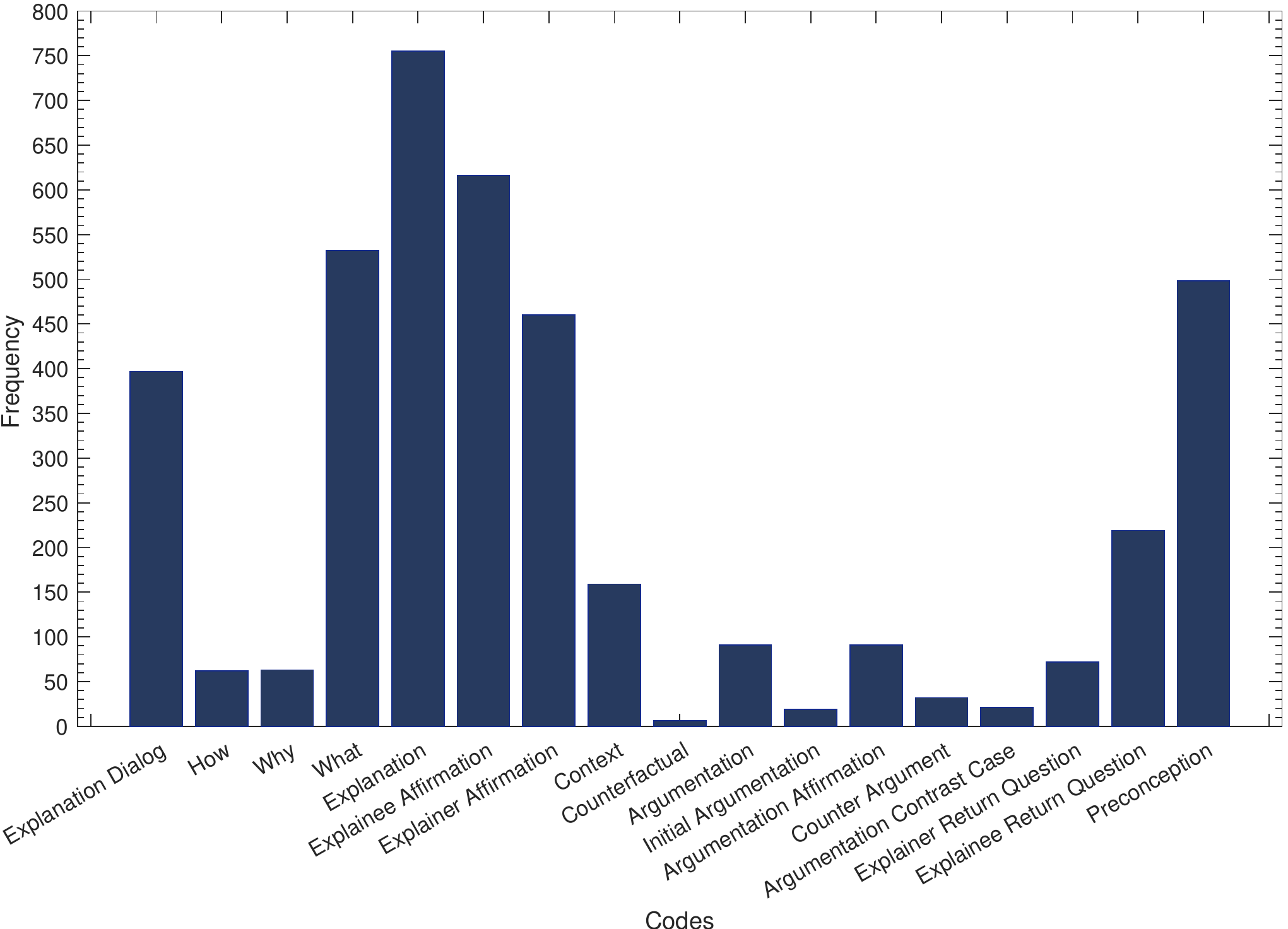}
\caption{Code Frequency.} \label{fig1}
\end{figure}

Figure \ref{fig1} shows the frequency of codes from all data sources. Across six different explanation types and 398 explanation dialogs, overall the most prevalent question type is `What' questions. Coding of `What' questions include all the other questions that are not categorized to `Why' and `How' questions and include questions types such as Where, Which, Do, Is etc. `Explanation' coding type has the highest frequency, which suggests it is likely to occur multiple times in a single explanation dialog. 

\begin{figure}[!h]
\includegraphics[width=\textwidth]{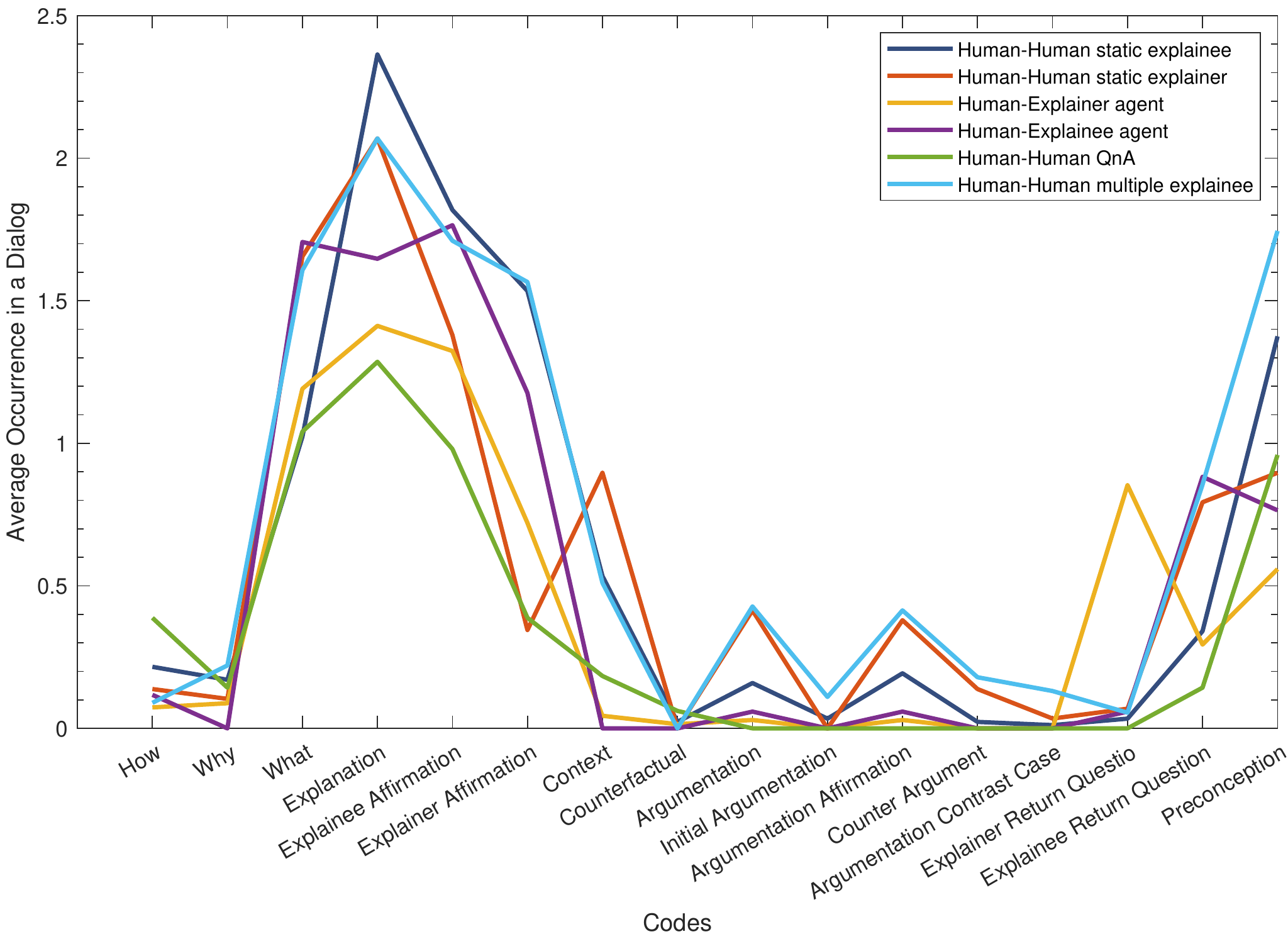}
\caption{Average code occurrence in per dialog in different explanation dialog types} \label{fig2}
\end{figure}

The average code occurrence per dialog in different dialog types is depicted in Figure \ref{fig2}. According to Figure \ref{fig2} in all dialog types, a dialog is most likely to have multiple what questions, multiple explanations and multiple affirmations. Humans tend to provide context and further information surrounding a question, whereas Human-Agent dialogs almost never initiate questions with context. This could be due to the humans prejudice of the incapability of the agent to identify the given context. In contrast, Human-Explainer agent dialog types have explainee return questions present more than Human-Human dialogs. Further analysis of the data revealed this is due to nature of the chatbot the data source was based upon, where the chatbot try to hide its weaknesses by repeatedly asking unrelated questions.  Human-only dialogs also have a higher frequency of `Preconception' occurrences. We attribute this to the nature of Human-Human explanations where humans try to express their opinion or preconceived views.

Argumentation is a key component of an explanation dialog. The explainee can have different or contrasting views to the explainer regarding the explanation, at which point an argument can be put forth by the explainee. An argument in the form of an explanation that is not in response to a question can also occur at the very beginning of an explanation dialog, where the argument set the premise for the rest of the dialog. An argument is typically followed by an affirmation and may include a counter argument by the opposing party. From Figure \ref{fig2} Human-Human dialogs with the exception of QnA have argumentation but Human-Agent dialogs lacks any substantial occurrences of argumentation. 

\subsubsection{Code Occurrence Analysis per Dialog}

\begin{figure}[!h]
\includegraphics[width=\textwidth]{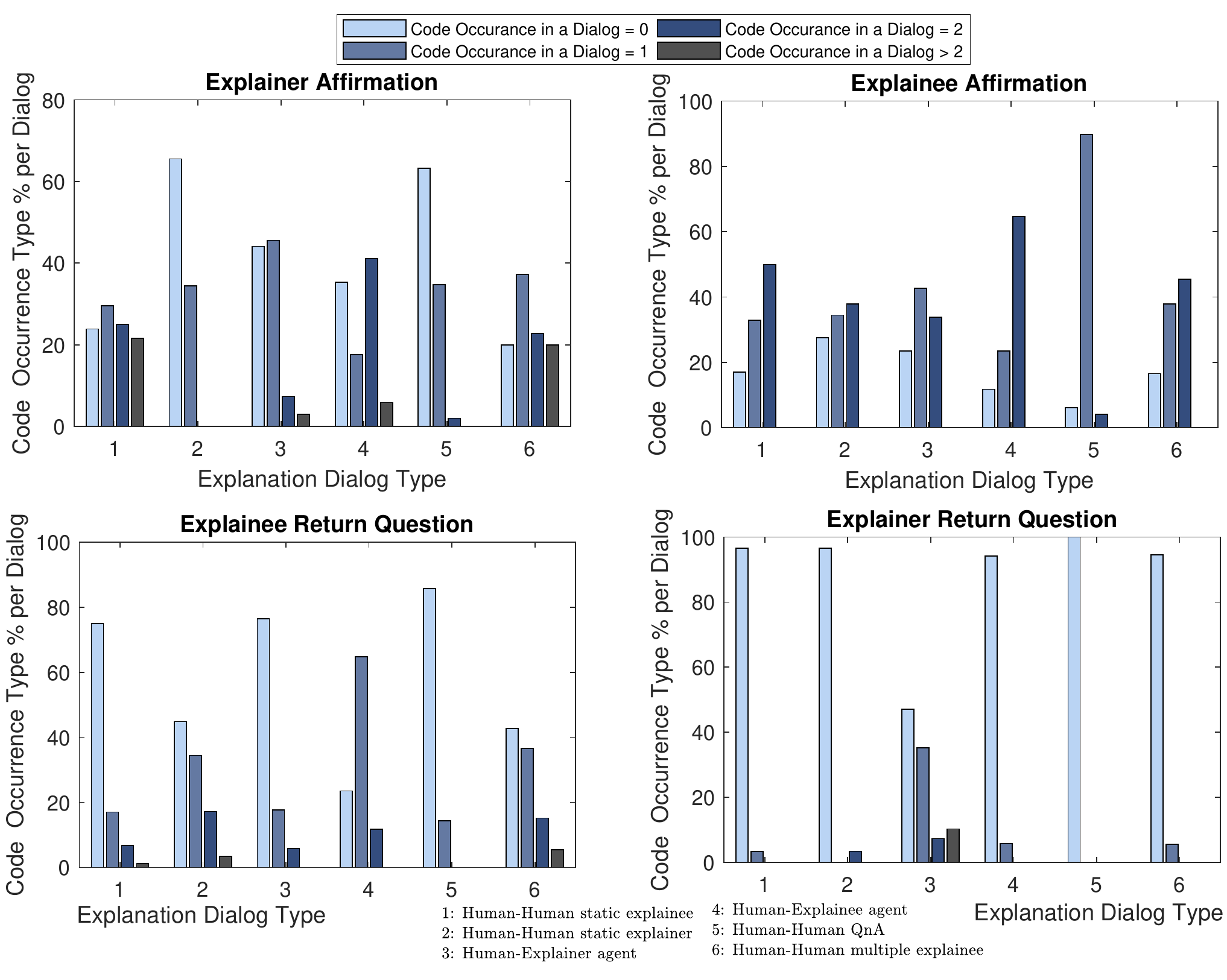}
\caption{Average code occurrence in per dialog in different explanation dialog types} \label{fig3}
\end{figure}

It is important to identify what is the most common occurrence frequency per dialog of certain codes in different dialog types. Analyzing common occurrences can help in identifying cyclic paths between components. For example, a majority of Human-Human static explainer dialogs have zero explainer affirmations, while Human-Human multiple explainer dialogs can have one explainer affirmation per dialog for majority of the dialogs. However, it is important to note that non-verbal cues were not in the transcripts, so non-verbal affirmations such as nodding were not coded. 

Figure~\ref{fig3} illustrate the occurrence frequency and the likelihood of four codes, and the above example can be seen in the explainer affirmation plot. These frequencies can be used to determine the sequence and cycles of components when formulating an explanation dialog model. For example, from Figure \ref{fig3}, the path or the sequence of a dialog model that have explainer return question is the least likely to occur as all explanation dialog types have zero explainer return question as the most probable one.

\subsubsection{Explanation Dialog Ending Sequence Analysis}

Participants should be able to identify when an explanation dialog ends. We analyse the different types of explanation dialogs to identify the codes that are most likely to signify the ending.

\begin{figure}[!h]
\includegraphics[width=\textwidth]{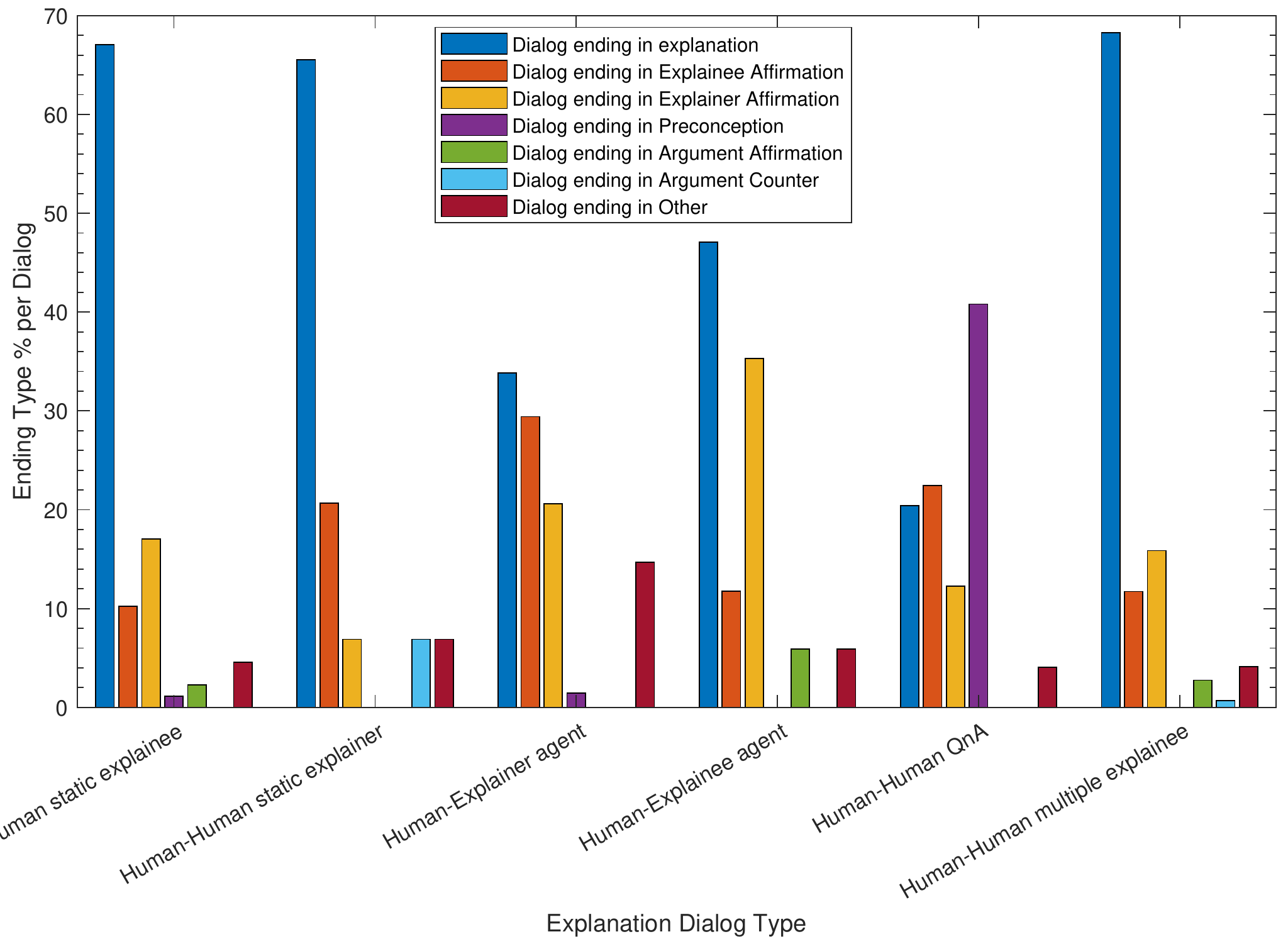}
\caption{Average code occurrence in per dialog in different explanation dialog types} \label{fig4}
\end{figure}

From Figure \ref{fig4}, all explanation dialog types except Human-Human QnA type are most likely to end in an explanation. The second most likely code to end an explanation is explainer affirmation. Ending with other codes such as explainee and explainer return questions is presented by `Dialog ending In Other' bar in Figure \ref{fig4}. It is important to note that although a dialog is likely to end in an explanation, that dialog can have previous explainee affirmations and explainer affirmations.

\section{Discussion}

The purpose of this study is to introduce an Explanation Dialog model that can bridge the gap of communicating explanations that are given by Explainable Artificial intelligence (XAI) systems. We derive the model from components identified through conducting a grounded study. The study consisted of coding data of six different types of explanation dialogs from six data sources.

We believe our proposed model can be used as a basis for building communication models for XAI systems. As the model is developed through analysing explanation dialogs of various types, a system that has the model implemented can potentially carry out an explanation dialog that can closely emulate a Human explanation dialog. This will enable the XAI systems to better communicate the underlying explanations to the interacting humans.

One key limitation of our model is its inability to evaluate effectiveness of the delivered explanation. In natural explanation dialogs that we analysed this evaluation did not occur, although we note that other data sources, such as verbal examinations or teaching sessions, would likely change these results. Although the model does not have a hard explanation evaluation system, it does accommodate explanation acknowledgement in the form of affirmation where the explanation evaluations can be embedded in the affirmation. In this case, further processing of the affirmation may be required. A key limitation if our study in general is the lack of empirical study to evaluate the suitability of model in a human-agent explanation. The proposed model is based on text based transcribed data sources, where an empirical study can introduce new components or relations; for example, interactive visual explanations.

\section{Conclusion}

Explainable Artificial Intelligent systems can benefit from having a proper explanation model that can explain their actions and behaviours to the interacting users. Explanation naturally occurs as a continuous and iterative socio-cognitive process that involves two processes: a cognitive process and a social process. Most prior work is focused on providing explanations without sufficient attention to the needs of the explainee, which reduces the usefulness of the explanation to the end-user.

In this paper, we propose a dialog model for the socio-cognitive process of explanation. Our explanation dialog model is derived from different types of natural conversations between humans as well as humans and agents. We formulate the model by analysing the frequency of occurrences of patterns to identify the key components that makeup an explanation dialog. We also identify the relationships between components and their sequence of occurrence inside a dialog. We believe this model has the ability to accurately emulate the structure of an explanation dialog similar to a one that occurs naturally between humans. Socio-cognitive systems that deal in explanation and trust will benefit from such a model in providing better, more intuitive and interactive explanations. We hope XAI systems can build on top of this explanation dialog model and so lead to better explanations to the intended user.

In future work, we aim to evaluate the model in a Human-Agent setting. Further evaluation can be done by introducing other forms of interaction modes such as visual interactions which may introduce different forms of the components in the model.

\section{Acknowledgments}

The research described in this paper was supported by the University of Melbourne research scholarship (MRS); SocialNUI: Microsoft Research Centre for Social Natural User Interfaces at the University of Melbourne; and a Sponsored Research Collaboration grant from the Commonwealth of Australia Defence Science and Technology Group and the Defence Science Institute, an initiative of the State Government of Victoria.

%
%
%
%
\bibliographystyle{splncs04}
\bibliography{main}

\end{document}